\ifdefined\XeTeXversion

\RequirePackage[T1]{fontenc}
\fi
\documentclass[letterpaper,10pt,conference]{ieeeconf}

\IEEEoverridecommandlockouts
\usepackage{graphicx}
\usepackage{xcolor}
\newcommand{\revised}[1]{\textcolor{black}{#1}}
\usepackage{hyperref}
\usepackage{amsmath,amssymb}
\usepackage{booktabs}
\usepackage{multirow}

\usepackage[ruled,vlined]{algorithm2e}

\title{Stubborn: A Streamlined and Unified Reinforcement Learning Framework for Robust Motion Tracking and Fall Recovery for Humanoids}

\author{Yuhui~Yang\textsuperscript{\footnotesize\ensuremath{\dagger}}, 
Xiao~Ren\textsuperscript{\footnotesize\ensuremath{\dagger}}, Zongbiao~Weng\textsuperscript{\footnotesize}, Zhijie~Liu\textsuperscript{\footnotesize}, and He~Kong\textsuperscript{\footnotesize\ensuremath{*}}%
\thanks{The authors are with the School of Automation and Intelligent Manufacturing, Southern University of Science and Technology, Shenzhen 518055, China. Emails: [12411024;12431359;12433032]@mail.sustech.edu.cn, liuzhijie2000@foxmail.com, kongh@sustech.edu.cn.
}%
\thanks{\authorrefmark{2}Y.~Yang and X.~Ren contributed equally to this work.}
\thanks{\authorrefmark{1}Corresponding author. This work was supported in part by the National Natural Science Foundation of China under Grant 62550003, in part by the Science, Technology, and Innovation Commission of Shenzhen Municipality, China under Grant JCY20240813094212017, in
part by the Shenzhen Science and Technology Program under Grant KQTD20221101093557010, in part by Guangdong Science and Technology Program under Grant 2024B1212010002.}}%
\hypersetup{pdfauthor={Yuhui Yang, Xiao Ren, Zongbiao Weng, Zhijie Liu, He Kong}}

\begin{document}
\maketitle

%===============================================================================

\begin{abstract}
% Robust motion tracking and autonomous fall recovery are important for the practical deployment of humanoid robots. 
Recent reinforcement learning approaches have shown great promise in improving humanoid motion tracking performance and achieving fall recovery under disturbances. However, most existing works treat motion tracking and fall recovery as different tasks and require multi-stage training with specialized recovery rewards and/or separate recovery policies. Moreover, existing reinforcement learning-based methods often terminate training episodes immediately after severe tracking failures, limiting recovery-oriented exploration in unstable or fallen states.
% recovery from disturbances and falls remain challenging problems in humanoid motion control under highly dynamic motions and strong external perturbations.
% Existing reinforcement learning-based motion tracking methods often terminate training episodes immediately after severe tracking failures, limiting recovery-oriented exploration in unstable or fallen states.
% As a result, policies often struggle to recover tracking after strong disturbances or a complete loss of balance.
To address the above issues, we propose Stubborn, a streamlined and unified reinforcement learning framework to achieve robust humanoid motion tracking and fall recovery.  
% The framework employs an asymmetric Actor-Critic architecture during training to utilize privileged information from the simulation environment, while relying only on proprioceptive observations and reference motion features during deployment.
Specifically, Stubborn uses an asymmetric Actor-Critic architecture and consists of three major components. First, a yaw-aligned tracking representation is adopted to reduce sensitivity to global drift and heading disturbances while preserving gravity-related balance information. Second, 
% in contrast to existing works that terminate training episodes immediately after severe tracking failures, 
we introduce a Bernoulli-based probabilistic termination mechanism that enables the policy to encourage exploration of fall-recovery behaviors under varying failure modes.
% explore disturbance/fall recovery behaviors after severe tracking failures. 
% to encourage continuous exploration of fall-recovery dynamics under disturbance
Third, we propose a probabilistic termination and tracking-error-driven strategy that dynamically reshapes the sampling distribution based on tracking performance, increasing the training efficiency for difficult motion segments and unstable states. Extensive comparisons with SOTA methods and ablation studies show that Stubborn achieved competitive performance \revised{and avoided the reward-hacking tendency of baseline methods}, and the proposed probabilistic termination mechanism and adaptive sampling strategy contributed to the performance and robustness gains. For real-world demonstrations, please refer to the project page at https://aislab-sustech.github.io/Stubborn/.
% {https://aislab-sustech.github.io/RoMTrack/}
\end{abstract}

% {\small
% \noindent\revised{\textbf{Project page:}\\
% \url{https://aislab-sustech.github.io/Stubborn/}}\par
% }

% Keywords (not shown in the ICRA review manuscript): Humanoids, reinforcement learning, motion tracking, fall recovery
\section{Introduction}

Recent years have witnessed much progress in reinforcement learning-based highly dynamic motion tracking for humanoids, including dancing~\cite{peng2021amp, luo2023perpetual, sun2025robotdancing} and martial arts~\cite{xie2025kungfubot, han2025kungfubot2}, and crawling~\cite{xu2026reach}. There exist works that rely on global absolute pose tracking for training, which might cause the policy to exert excessive control effort to correct translation and heading deviations after perturbations~\cite{peng2018deepmimic, zhu2026clot}. To address the above issue, recent works in the literature adopt a yaw-aligned tracking representation to reduce sensitivity to global drift and heading disturbances while preserving gravity-related balance information~\cite{cheng2025rambo, lee2025learning}.

% Therefore, we will adopt a yaw-aligned tracking representation in this paper.

Although existing reinforcement learning (RL) methods can achieve remarkable performance under normal conditions, maintaining tracking accuracy and robustness (e.g., recovering from falls) under strong external disturbances remains a challenging problem~\cite{yin2025unitracker, wu2025learn, jung2025ppf}. In particular, how to enhance tracking performance and robustness in an effective and unified way, rather than treating motion tracking and fall recovery as separate tasks, has not been well explored and several design issues remain to be addressed. 

% several issues still remain regarding tracking error representation, the design of termination mechanisms, and, more importantly, the enhancement of motion-tracking performance and robustness in an effective and unified way.

% Current approaches often terminate training episodes immediately after severe tracking failures, limiting recovery-oriented exploration in unstable or fallen states.
% Despite recent progress, robust tracking of highly dynamic motions remains challenging for humanoid robots under external disturbances and unstable conditions~\cite{yin2025unitracker, wu2025learn, jung2025ppf}.
% In particular, under long-horizon training, uniformly sampled training data often under-represent difficult motion segments and unstable states, limiting the exploration capability of the policy~\cite{liao2025beyondmimic, chen2025gmt}. In addition, for fall recovery, most existing methods employ hard termination mechanisms: once the tracking error exceeds a predefined threshold, the policy training terminates (or, during deployment, the execution terminates). Such designs would prematurely prevent the policy from exploring unstable or fallen states and the corresponding action space to learn to recover, making this type of behavior difficult to acquire.

In particular, under long-horizon training, uniformly sampled training data often under-represent difficult motion segments and unstable states, limiting the exploration capability of the policy~\cite{liao2025beyondmimic, chen2025gmt}. For fall recovery, most existing methods employ hard termination mechanisms: once the tracking error exceeds a predefined threshold, the policy training terminates (or, during deployment, the execution terminates). Such designs would prematurely prevent the policy from exploring and learning to recover.
% truncate interaction in unstable or fallen states, making recovery behaviors difficult to acquire.
Existing studies often address this issue through recovery-specific rewards, multi-stage training procedures, or separate recovery controllers~\cite{duburcq2022reactive, yang2025bracing}, thus increasing the architectural or training complexity. 

% In addition, under long-horizon training, uniformly sampled training data often under-represent difficult motion segments and unstable states, limiting tracking robustness and recovery performance.

%%%%%%%%%%%%%%%%%%%%%%%%%%%%%%%%%%%%%%%%%%%%%%%%%%%%%%%%%%%%%%%%%%%%%%%%%%%%%%%%%%%%%%%%%%%%%%%%%%%%%%%%%%%%%%%%%%%%%%%%%%%%%%%
% %% Figure
% \begin{figure}[t!]
%     \centering
%     \includegraphics[width=\columnwidth]{Figures/tracking_recovery_overview.pdf}
%     \caption{ Our method enables highly dynamic motion tracking and fall recovery using one policy. A yaw-invariant tracking representation, a probabilistic termination mechanism, and a tracking-error-driven adaptive sampling strategy enable recovery from unstable states during complex dynamic motions.}
%     \label{fig:tracking_recovery_overview}
% \end{figure}
%%%%%%%%%%%%%%%%%%%%%%%%%%%%%%%%%%%%%%%%%%%%%%%%%%%%%%%%%%%%%%%%%%%%%%%%%%%%%%%%%%%%%%%%%%%%%%%%%%%%%%%%%%%%%%%%%%%%%%%%%%%%%%%

%% Figure
\begin{figure*}[t!]
    \centering
    \includegraphics[width=0.95\textwidth]{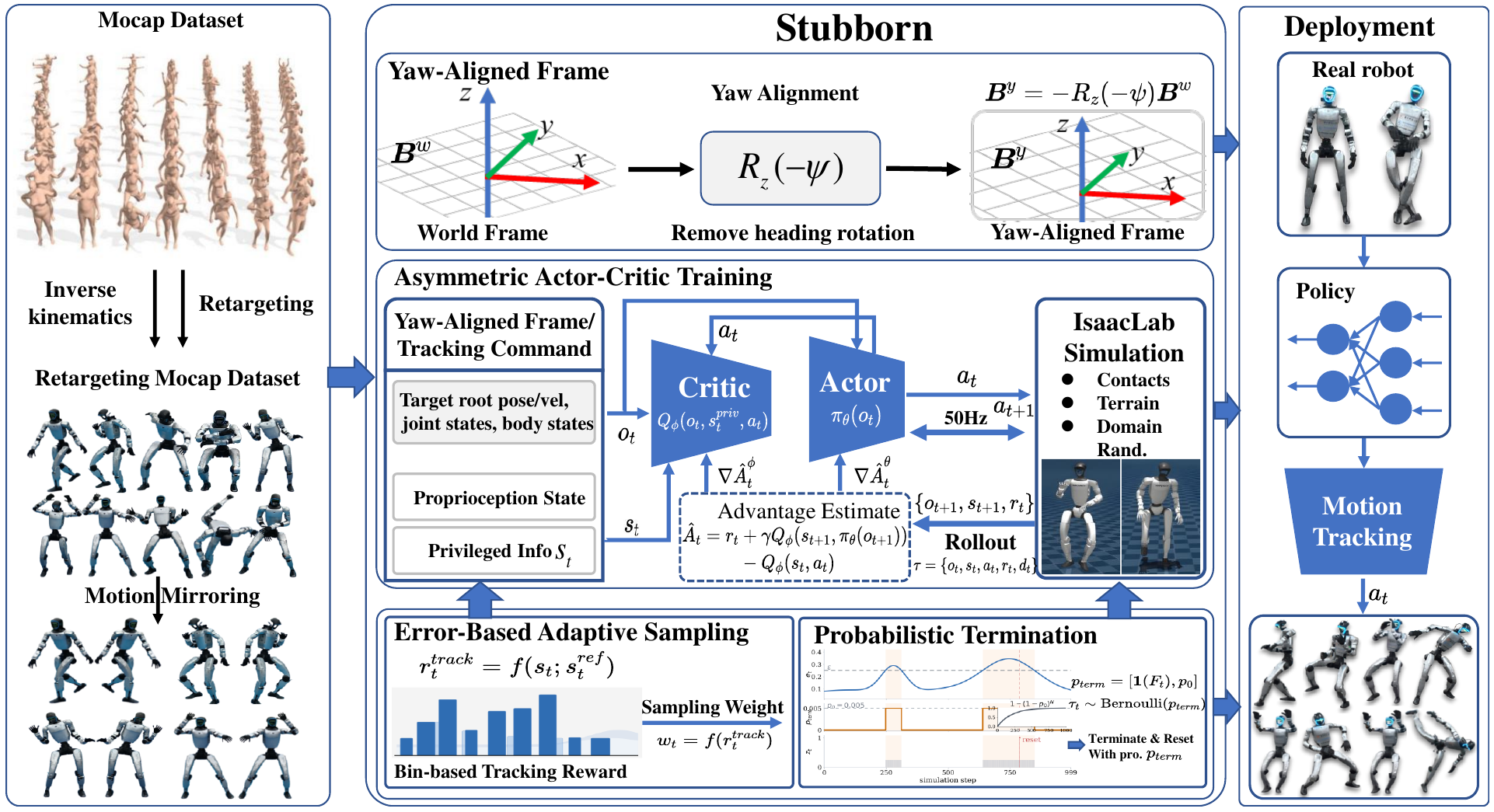}
    \caption{\textbf{Overview of Stubborn.} 
    % Motion-capture data are first retargeted into reference motions. 
    The policy is trained with a yaw-aligned representation and via an asymmetric actor-critic architecture, with a Bernoulli-based soft termination mechanism and tracking-error-based adaptive sampling.
    % and probabilistic termination. The learned policy is finally deployed on the real robot for highly dynamic motion tracking and recovery from unstable states.
    }
    \label{fig:framework}
\end{figure*}
%%%%%%%%%%%%%%%%%%%%%%%%%%%%%%%%%%%%%%%%%%%%%%%%%%%%%%%%%%%%%%%%%%%%%%%

To address these challenges, we propose Stubborn, a streamlined and unified RL framework for training a single policy to achieve robust humanoid motion tracking and fall recovery. As shown in Fig.~\ref{fig:framework}, Stubborn uses a unified single policy for tracking highly dynamic motions while maintaining robustness under strong perturbations and achieving recovery from fallen states. Stubborn adopts an asymmetric Actor-Critic architecture~\cite{radosavovic2024real} with yaw-aligned tracking representation, where the value network uses privileged simulation information, while the policy network relies only on proprioceptive observations.

On this basis, we make the following contributions: (1) We propose a Bernoulli-based soft termination mechanism that encourages the policy to learn resilient actions in unstable and fallen states, thereby synthesizing  emergent recovery behaviors without auxiliary recovery-specific reward design. (2) We develop a probabilistic termination (PT) and tracking error-driven sampling strategy that dynamically reshapes the sampling distribution based on online tracking performance, thereby improving learning efficiency in difficult tracking and recovery states. (3) Extensive simulations and real experiments on the 29-DoF Unitree G1 humanoid robot demonstrate that Stubborn achieves competitive performance against SOTA methods~\cite{HoloMotion,zhang2025track,li2025bfm,liao2025beyondmimic}. Ablation studies also show that the proposed PT mechanism and adaptive sampling strategy contribute to performance gains.

\section{Related Work}

\textbf{Humanoid Motion Tracking:} In recent years, reinforcement learning has become a common paradigm for humanoid motion imitation and tracking~\cite{peng2018deepmimic,liao2025beyondmimic,fu2024humanplus,he2024omnih2o}. 
For example, DeepMimic~\cite{peng2018deepmimic} learns physics-based controllers by tracking reference motion clips with imitation objectives. BeyondMimic~\cite{liao2025beyondmimic} learns motion-tracking primitives for highly dynamic humanoid skills through adaptive sampling and distills the learned primitives into a guided diffusion policy for downstream humanoid control. HumanPlus~\cite{fu2024humanplus} and OmniH2O~\cite{he2024omnih2o} primarily target whole-body imitation, teleoperation, and autonomous behavior learning, rather than unified motion tracking in highly dynamic, long-horizon settings. 

ASAP~\cite{he2025asap} improves sim-to-real transfer for agile humanoid skills through real-to-sim-to-real adaptation, while GMT~\cite{chen2025gmt} improves general motion tracking through large-scale motion data, adaptive sampling, and a motion mixture-of-experts architecture.
HoloMotion~\cite{HoloMotion} explores foundation models for whole-body  control to improve diverse motion generation and tracking, while BFM-Zero~\cite{li2025bfm} focuses on zero-shot and few-shot robot motion transfer. 
% providing a new direction for cross-morphology motion generalization. 

Motion tracking is closely related to whole-body control~\cite{liu2025opt2skill}, and data-driven whole-body control methods have further advanced this direction. 
ExBody2~\cite{cheng2024expressive,ji2024exbody2} emphasizes stable whole-body imitation under diverse reference motions. 
TWIST~\cite{ze2025twist,ze2025twist2} and CLONE~\cite{li2025clone} improve long-horizon whole-body control through real-time teleoperation and closed-loop correction, respectively. 
SONIC~\cite{luo2025sonic} further explores scaling motion tracking toward general whole-body behaviors. Although recent work continues to advance unified data-driven representations and broader generalization, challenges remain in robustness under strong perturbations, learning difficult long-horizon motion segments, and handling tracking and recovery within a unified framework.

\textbf{Robustness and Fall Recovery:} Motion robustness and autonomous fall recovery are important capabilities for humanoid robots in practice. RL-based methods have been widely used to train standing-up and push-recovery policies for humanoids~\cite{chen2025hifar, gaspard2025frasa, egle2024enhancing}. 
For example, HoST~\cite{huang2502learning} improves recovery across diverse poses through a multi-critic architecture and curriculum learning, while HumanUP~\cite{he2025learning} learns standing-up policies across multiple terrains and initial fall states. HuB~\cite{zhang2025hub} improves humanoid balance robustness by combining reference motion refinement, balance-aware policy learning, and robustness training. 

Although the above-mentioned methods help to learn recovery and balance behaviors, policies are typically designed for specific recovery or balance scenarios, separately from the motion tracking policy, requiring extensive reward tuning. 
% Beyond dedicated recovery policies, recent studies have explored disturbance-robust motion tracking.
Unlike recovery-specific policy learning, 
recent studies have also improved robustness from the perspective of motion tracking.
For example, Any2Track~\cite{zhang2025track} adopts a two-stage tracker-adapter framework to improve motion tracking robustness under disturbances and sim-to-real variations.
%considers stable tracking and online adaptation of diverse motions under a range of dynamic perturbations, including terrain changes, external forces, and variations in physical parameters.

% Although such methods improve robustness under perturbations, most existing studies still treat motion tracking and fall recovery as separate tasks and typically rely on dedicated recovery rewards, multi-stage training procedures, or separate recovery policies. 

% Hence, such methods improve robustness under perturbations, recovery from severe instability or fallen states within the same motion-tracking policy has received relatively limited attention. In contrast, many dedicated recovery studies rely on recovery-specific rewards, multi-stage training procedures, or separate recovery policies.

Overall, existing works often treat motion tracking and fall recovery as competing objectives with conflicting gradients~\cite{chen2025hifar}. To decouple these tasks, the community seeks solutions using multi-stage training and separate recovery policies with recovery-specific rewards.
% rely on recovery-specific rewards, multi-stage training procedures, or separate recovery policies. 
We argue that tracking failures actually provide a natural gateway to the fall state manifold. By harnessing these natural transitions rather than avoiding them, one might achieve high-fidelity motion tracking, robustness to strong perturbations, and recovery from unstable states with a unified training framework and policy. This is the objective that this paper aims to fulfill. Our framework spontaneously explores recovery behaviors alongside motion tracking, enabling highly sample-efficient, simultaneous exploration of both tasks within a single policy.

% The question of how to achieve high-fidelity motion tracking, robustness to strong perturbations, and recovery from unstable states, using a unified training framework and policy, remains an open problem. 

% This is the gap that this paper aims to fill. 

% Unlike the approaches discussed above, this work introduces a probabilistic termination mechanism within a unified motion-tracking framework to support recovery-related learning. This design reduces reliance on specialized recovery rewards and improves robustness under external perturbations.

%%%%%%%%%%%%%%%%%%%%%%%%%%%%%%%%%%%%%%%%%%%%%%%%%%%%%%%%%%%%%%%%%%%%%%%%%%%%%%%%%%%%%%%%%%%%%%%%%%%%%%%%%%%%%%%%%%%%%%%%%%%%%%%
% %% Figure
% \begin{figure*}[t!]
%     \centering
%     \includegraphics[width=\textwidth]{Figures/framework.pdf}
%     \caption{Overview of the Stubborn. Motion-capture data are first retargeted into reference motions. The policy is then trained with a yaw-aligned representation, an asymmetric actor--critic architecture, tracking-error-based adaptive sampling, and probabilistic termination. The learned policy is finally deployed on the real robot for highly dynamic motion tracking and recovery from unstable states.}
%     \label{fig:framework}
% \end{figure*}
% %%%%%%%%%%%%%%%%%%%%%%%%%%%%%%%%%%%%%%%%%%%%%%%%%%%%%%%%%%%%%%%%%%%%%%%

\section{Method}
Stubborn, as illustrated in Fig.~\ref{fig:framework}, is trained using an asymmetric actor-critic architecture and consists of three core modules.
% , i.e., drift-invariant yaw-aligned tracking representation, the Bernoulli-based PT mechanism, as well as the PT and tracking error-driven sampling strategy. 
Due to limited space, details of the drift-invariant yaw-aligned tracking representation are placed in the Appendix section while 
the Bernoulli-based PT mechanism, as well as the PT and tracking error-driven sampling strategy are described in the following. 
% Section~\ref{subsec:drift_invariant_yaw_aligned} introduces a drift-resistant tracking representation based on a yaw-aligned root frame, which reduces sensitivity to global drift. 
% Section~\ref{subsec:probabilistic_termination} presents a Bernoulli-based probabilistic termination mechanism that relaxes hard truncation in unstable states, thereby broadening the exploration space for recovery-oriented behaviors. 
% Section~\ref{subsec:adaptive_sampling} describes a tracking-error-driven adaptive sampling strategy, which dynamically adjusts the sampling distribution under probabilistic termination and improves tracking performance on long and complex motion sequences.

\subsection{Bernoulli-based Probabilistic Termination for Fall Recovery Learning}
\label{subsec:probabilistic_termination}

% \subsubsection{Conditional Bernoulli Termination Mechanism}
\textbf{Conditional Bernoulli Termination Mechanism:} In our framework, when the tracking error exceeds a predefined threshold, the environment does not terminate the episode deterministically; instead, it triggers a termination with a user-specified probability. More specifically, we denote the height error $e_{root,z}^p$ of the robot's floating base and the orientation error $e_{root}^q$, computed via the quaternion inner product in SO(3), as follows:
\begin{equation}
\begin{aligned}
    e_{root,z}^p &= p_{root,z}^{ref} - p_{root,z}, \\
    e_{root}^q   &= 2 \arccos\left( \left| \langle q_{root}^{ref}, q_{root} \rangle \right| \right).
\end{aligned}
\end{equation}
Let $\tau_t \in \{0, 1\}$ denote the binary termination indicator at time step $t$. We model this indicator via a conditional Bernoulli distribution:
\begin{equation}
    P(\tau_t = 1 \mid s_t) = 
    \begin{cases} 
        p_{term}, & \begin{aligned}[t]
            &(|e_{root,z}^p| > \theta_{pos}) \\
            &{}\lor (e_{root}^q > \theta_{quat})
        \end{aligned} \\
        0, & \text{otherwise} 
    \end{cases} ,
\end{equation}
where $s_t$ represents the system state. \revised{The termination probability $p_{term}$ is calibrated to provide an expected exploration horizon compatible with the physical time required for fall recovery.} In the following, we provide guidelines for selecting $p_{term}$.

% \pi/2 \text{rad}$.

% \subsubsection{Design of the Recovery Time Window}
\textbf{Design of the Termination Probability:} If the robot remains in a marginally stable state after exceeding the tracking error threshold, the number of additional survival steps before episode termination can be modeled as a random variable $\mathcal{T} \sim Geo(p_{term})$. 
To maintain consistent temporal difference (TD) updates, the expectation of $\mathcal{T}$ should align with the physical recovery time:
\begin{equation}
    \mathbb{E}[\mathcal{T}] = \frac{1}{p_{term}} = f_{ctrl} \cdot t_{rec},
\end{equation}
where $f_{ctrl} = 50 \text{ Hz}$ is the control frequency. We set the physical recovery time to $t_{rec} = 4 \text{ s}$, which provides the policy with an expected exploration horizon of $200$ steps ($\mathbb{E}[\mathcal{T}] = 200$). 
A window that is too restrictive can degenerate the mechanism into a deterministic one, thus hindering the learning of recovery behaviors, and vice versa.
% Conversely, an excessively large window introduces inefficient sampling and value estimation bias following an irreversible fall. 
Following this design guideline, we select $p_{term}=0.005$, and set the thresholds to $\theta_{pos}=0.25\,\mathrm{m}$ and $\theta_{quat}=\frac{\pi}{2}\,\mathrm{rad}$ in our experiments. We additionally evaluate the sensitivity to $p_{term}$ in Section~\ref{sec:result}, including the degenerate case $p_{term}=0$.

% Under this mechanism, the survival probability of the policy decays geometrically: $P(\mathcal{T} > k)=(1-p_{term})^k$. While suppressing divergent trajectories, this characteristic expands the exploration boundary of marginally stable states, enabling the policy to autonomously learn recovery behaviors without explicit rewards.
\revised{Under this mechanism, at each step at which a termination threshold is exceeded, the episode continues with probability $1-p_{term}$; once both errors return within their limits, threshold-triggered termination is inactive.}
\revised{For consecutive threshold violations, let $k$ denote the number of Bernoulli tests; then the survival probability is $P(\mathcal{T}>k)=(1-p_{term})^k$.}
This design reduces immediate termination after disturbances or falls and provides a probabilistic recovery window for learning recovery behaviors toward post-perturbation stable tracking.
Furthermore, assuming a perturbation occurs early in the episode, under an ideal exploration window with a maximum episode length ($T_{\max} = 1000$), the cumulative probability of terminating prior to the step limit is given by $P(\mathcal{T} \le T_{\max}) = 1 - (1 - p_{term})^{T_{\max}}.$
% \begin{equation}
%     P(\mathcal{T} \le T_{\max}) = 1 - (1 - p_{term})^{T_{\max}}.
% \end{equation}
% \revised{Substituting our selected parameters ($p_{term}=0.005$, $T_{\max}=1000$), this probability evaluates to approximately $0.993$, conditional on violating a termination threshold throughout all $1000$ steps. This calculation describes persistently unstable trajectories; recovered trajectories are no longer subject to these Bernoulli tests.}

\subsection{Probabilistic Termination and Tracking Error Driven Sampling}
\label{subsec:adaptive_sampling}
% Many motion-tracking methods employ adaptive sampling strategies coupled with hard termination mechanisms, increasing the sampling probability of difficult segments based on motion completion, failure rate, or episode termination frequency~\cite{liao2025beyondmimic}. 
Under hard termination, segments with tracking errors exceeding a predefined threshold are more likely to terminate the episode early~\cite{liao2025beyondmimic}. However, under the PT mechanism described in Section~\ref{subsec:probabilistic_termination}, segments exceeding the error threshold are not terminated immediately; instead, the episode terminates only with a certain probability.
% As a result, episode termination is no longer determined solely by segment difficulty, but also by motion complexity, external perturbations, and stochastic termination.

As a result, episode termination is no longer a reliable proxy for segment difficulty, since it is affected by stochastic termination and external perturbations.
Consequently, the correspondence between termination statistics and segment difficulty is weakened, making hard-sample mining based on termination statistics less reliable.
To address this issue, we propose a probabilistic termination and tracking-error-driven adaptive sampling strategy.

Let the total length of the reference trajectory be $N$, and let $w_t$ denote the unnormalized sampling weight associated with reference frame $t$. 
At each environment reset, an initial frame $t_0$ is sampled according to the normalized sampling distribution, and an episode of maximum length $T_{\max}$ is executed from that frame.
Let $T$ denote the actual episode length, where $1 \le T \le T_{\max}$, and let $e_{\mathrm{kp},\,t}$ denote the instantaneous keypoint tracking error at reference frame $t$. 
Then, for an episode starting at $t_0$ and lasting $T$ steps, the average keypoint tracking error is defined as
\begin{equation}
    \bar{e}=\frac{1}{T}\sum_{k=0}^{T-1} e_{\mathrm{kp},\,t_0+k},
\end{equation}
where $\bar{e}$ represents the average keypoint tracking error over the current reference segment. For each time step $t \in [t_0,\, t_0+T-1]$ within the sampled segment, the weight increment $\Delta w_t$ is defined as
\begin{equation}
    \Delta w_t=
    \begin{cases}
        +\Delta w_i, & \bar{e}\ge \theta_{\text{success}},\\[4pt]
        \Phi_{\text{att}}(t), & \bar{e}<\theta_{\text{success}} \ \land\ T=T_{\max},\\[4pt]
        \Phi_{\text{dist}}(t), & \bar{e}<\theta_{\text{success}} \ \land\ T<T_{\max}.
    \end{cases}
\end{equation}
Here, $\theta_{\text{success}}$ denotes the success threshold for the average tracking error, and we set $\theta_{\text{success}}=0.06$ in our implementation. 
A segment is regarded as successfully tracked when $\bar{e}<\theta_{\text{success}}$; otherwise, it is considered to require further training. 
Moreover, $\Delta w_i>0$ and $\Delta w_d>0$ denote the step sizes for increasing and decreasing the sampling weights, respectively.
In our implementation, $\Delta w_i=0.10$ and $\Delta w_d=0.05$. 

When an episode reaches the maximum length $T_{\max}$ while maintaining a low average error, the local decay operator $\Phi_{\text{att}}(t)$ is applied. 
When an episode satisfies $\bar{e}<\theta_{\text{success}}$ but terminates before reaching the maximum length, i.e., $T<T_{\max}$, the local redistribution operator $\Phi_{\text{dist}}(t)$ is applied. 
$\Phi_{\text{att}}(t)$ and $\Phi_{\text{dist}}(t)$ are defined as 
% \begin{equation}
%     \begin{aligned}
%         \Phi_{\text{att}}(t) &=
%         \begin{cases}
%             -2\Delta w_d, & t<t_0+\lfloor T/2 \rfloor,\\
%             -\Delta w_d, & \text{otherwise},
%             \end{cases}\\[6pt]
%             \Phi_{\text{dist}}(t) &=
%             \begin{cases}
%             -2\Delta w_d, & t<t_0+\lfloor T/3 \rfloor,\\
%             -\Delta w_d, & t<t_0+\lfloor T/2 \rfloor,\\
%             0, & t<t_0+\lfloor 2T/3 \rfloor,\\
%             +\Delta w_i, & \text{otherwise}.
%         \end{cases}
%     \end{aligned}
% \end{equation}
\begin{equation}
\begin{aligned}
\Phi_{\text{att}}(t)
&=
\begin{cases}
-2\Delta w_d, & t<t_0+\lfloor T/2 \rfloor, \\
-\Delta w_d, & \text{otherwise},
\end{cases}
\\[4pt]
\Phi_{\text{dist}}(t)
&=
\begin{cases}
-2\Delta w_d, & t<t_0+\lfloor T/3 \rfloor, \\
-\Delta w_d, & t<t_0+\lfloor T/2 \rfloor, \\
0, & t<t_0+\lfloor 2T/3 \rfloor, \\
+\Delta w_i, & \text{otherwise}.
\end{cases}
\end{aligned}
\end{equation}

Subsequently, the sampling weights of the reference frames involved in the current episode are updated as follows:
where $w_{\min}$ and $w_{\max}$ denote the lower and upper bounds of the sampling weights, respectively. 
In our implementation, we set $w_{\min}=0.05$ and $w_{\max}=1.0$. 
After the update, the weights are normalized to obtain the sampling distribution for selecting the starting frame of the next episode as follows 
% $p_t=\frac{\max(w_t,w_{\min})}{\sum_{j=0}^{N-1}\max(w_j,w_{\min})}$.
\begin{equation*}
     p_t=\frac{\max(w_t,w_{\min})}{\sum_{j=0}^{N-1}\max(w_j,w_{\min})}.
\end{equation*}
In this way, the online tracking error $\bar{e}$ serves as an empirical indicator of segment difficulty, allowing the sampling distribution to adapt dynamically even under the probabilistic termination mechanism.
As a result, more probability mass is assigned to high-error regions, encouraging training to focus on difficult motion segments. 
\revised{The overall algorithm of Stubborn is summarized in Algorithm~\ref{alg:whole_body_tracking_recovery}.}

\begin{algorithm}[!t]
\small
\caption{The Overall Algorithm Pipeline of Stubborn}
\label{alg:whole_body_tracking_recovery}

\SetKwInOut{Input}{Input}
\SetKwInOut{Output}{Output}

\Input{
    \begin{tabular}[t]{@{}l @{\hspace{0.3em}:\hspace{0.3em}} l@{}}
        $\mathcal{T}_{ref} \in SE(3)^{N \times |\mathcal{B}|}$ & Reference trajectory \\
        $\pi_\theta, V_\phi$ & Policy and Value networks \\
        $N_{env}, K_{rollout}$ & Env count \& rollout length \\
        $\begin{gathered}\theta_{success}, \theta_{pos},\\\theta_{quat}, p_{term}\end{gathered}$ & Threshold parameters
    \end{tabular}
}
\Output{
    \begin{tabular}[t]{@{}l @{\hspace{0.5em}:\hspace{0.5em}} l@{}}
        $\theta, \phi$ & Optimized network parameters
    \end{tabular}
}
\vspace{0.3em}

Init $w_t \gets 1.0, \forall t$; Sample states $s^{(e)} \sim p_t \propto w_t$\;

\For{\text{iter} $= 1, 2, \dots$}
{
    \For{$k = 1$ \KwTo $K$}
    {
        $a_k^{(e)} \sim \pi_\theta(\cdot|s_k^{(e)}), \;\; s_{k+1}^{(e)} \gets \text{EnvStep}(s_k, a_k)$\;

        \tcp{\textbf{Drift-Invariant Yaw-Aligned Tracking Representation}}
        $T_\Psi^W \gets \text{YawAlign}(T_B^W)$ \;
        $(\tilde{p}_i, \tilde{R}_i)^{(e)} \gets (T_\Psi^W)^{-1} T_i^W, \;\forall i \in \mathcal{B}$\;
        $r_k^{(e)} \gets \text{Reward}(\{\tilde{p}_i, \tilde{R}_i\})$\;

        \tcp{\textbf{Bernoulli-based Probabilistic Termination for Fall Recovery Learning}}
        \For{env $e \in \{1, \dots, N_{env}\}$}
        {
            $\rho_k \gets \mathbb{I}\big( |e_{root,z}^p| \!>\! \theta_{pos} \lor e_{root}^q \!>\! \theta_{quat} \big)$\;
            $\tau_k \sim \text{Bernoulli}(\rho_k \cdot p_{term})$\;

            \If{$T \ge T_{\max} \lor \tau_k == 1$}{
                \tcp{\textbf{Probabilistic Termination and Tracking Error-driven Sampling}}
                $\bar{e} \gets \text{MeanErr}(t_0^{(e)}, T)$\;
                \For{$t \in [t_0^{(e)}, t_0^{(e)} + T - 1]$}
                {
                    \lIf{$\bar{e} \ge \theta_{success}$}{$\Delta w_t \gets +\Delta w_i$}
                    \lElseIf{$T \ge T_{\max}$}{$\Delta w_t \gets \Phi_{att}(t)$}
                    \lElse{$\Delta w_t \gets \Phi_{dist}(t)$}

                    $w_t \gets \max(\min(w_t \!+\! \Delta w_t, w_{max}), w_{min})$\;
                }
                $p_t \gets w_t/\sum w_j, \quad t_0^{(e)} \sim \text{Cat}(p_t)$\;
                Reset $s^{(e)}$ to ref. pose $t_0^{(e)}$ + perturbs\;
            }
        }
    }

    \tcp{\textbf{Policy Optimization}}
    Estimate GAE $\hat{A}$ using $\{s, a, r\}_{1:K}^{(1:N_{env})}$\;
    $\theta, \phi \gets \arg\max_{\theta, \phi} \mathbb{E} \big[ \mathcal{L}_{clip} - c_v \mathcal{L}_{VF} \big]$\;
}
\Return $\theta, \phi$\;
\end{algorithm}

\section{Results and Discussions}
\label{sec:result}
This section provides a systematic evaluation of Stubborn through both simulation and real experiments, with a focus on the following four questions: \textbf{Q1}: How does Stubborn compare with baseline methods on large and diverse motion datasets, and how well does it generalize to unseen motions? \textbf{Q2}: \revised{How does PT affect disturbance and fall recovery, how does it perform compared to a fixed recovery window perform, how sensitive is the learned policy to the termination probability parameter $p_{term}$?} \textbf{Q3}: Does the proposed PT and tracking error-driven sampling strategy improve tracking performance across diverse motion skills? \textbf{Q4}: How does Stubborn perform in real-world scenarios?

% \begin{itemize}
%     \item \textbf{Q1}: How does Stubborn compare with baseline methods on large and diverse motion datasets, and how well does it generalize to unseen motions?
%     \item \textbf{Q2}: Does the PT mechanism improve disturbance recovery and fall recovery capability under strong perturbations?
%     \item \textbf{Q3}: Does the proposed PT and tracking error-driven sampling strategy improve tracking performance across diverse motion skills?
%     \item \textbf{Q4}: How does Stubborn perform in real-world scenarios?
% \end{itemize}

\subsection{ Experiment Setup}
% This paper evaluates the motion-tracking performance of Stubborn on both simulation and real-robot platforms.
% We evaluate Stubborn on both simulation and real-robot platforms.
% Simulation experiments are conducted in IsaacLab/MuJoCo using motions from the public LAFAN1~\cite{harvey2020robust} dataset.
% Real-robot experiments are performed on the 29-DoF Unitree G1 humanoid robot using retargeted motions from both LAFAN1~\cite{harvey2020robust} and AMASS~\cite{mahmood2019amass} to evaluate the practical deployability of the learned policy.
% This paper evaluates the motion-tracking performance of Stubborn on both simulation and real-robot platforms. 
Simulation experiments are conducted in IsaacLab/MuJoCo using motions from the public LAFAN1~\cite{harvey2020robust} dataset. 
Real-robot experiments are performed on the 29-DoF Unitree G1 humanoid robot with retargeted motions from LAFAN1~\cite{harvey2020robust} and AMASS~\cite{mahmood2019amass} to evaluate the practical deployability of the learned policy.

% \textbf{Baselines.} We compare Stubborn with several representative humanoid motion tracking methods, including HoloMotion~\cite{HoloMotion}, Any2Track~\cite{zhang2025track}, BFM-Zero~\cite{li2025bfm}, and a From-scratch multi-motion RL baseline~\cite{wang2026omnixtreme}. Specifically, this baseline follows the algorithmic framework and network architecture of BeyondMimic~\cite{liao2025beyondmimic}, while extending single-motion tracking to an end-to-end reinforcement learning setting for multi-motion training. This comparison aims to evaluate Stubborn's performance in motion-tracking tasks.
% We do not directly compare with OmniXtreme~\cite{wang2026omnixtreme} because it is trained with substantially larger-scale motion data and targets general high-dynamic humanoid control, whereas our evaluation focuses on precise motion tracking, robustness, and fall recovery.
\textbf{Baselines.} We compare Stubborn with representative humanoid motion tracking methods, including HoloMotion~\cite{HoloMotion}, Any2Track~\cite{zhang2025track}, BFM-Zero~\cite{li2025bfm}, and a From-scratch multi-motion RL baseline~\cite{wang2026omnixtreme}. The latter follows BeyondMimic~\cite{liao2025beyondmimic} in its algorithmic framework and network architecture, while extending single-motion tracking to end-to-end multi-motion RL training. We do not directly compare with OmniXtreme~\cite{wang2026omnixtreme}, as it uses substantially larger-scale motion data and targets general highly dynamic humanoid control, whereas our evaluation focuses on precise motion tracking, robustness, and fall recovery.

% \textbf{Evaluation Metrics.} We evaluate policy performance in terms of motion tracking and fall recovery.
% For the motion tracking task, the root-relative mean per-body position error (MPBPE, mm), mean per-joint position error (MPJPE, $10^{-3}$ rad), mean per-joint velocity error (MPJVE, $10^{-3}$ rad/s), and acceleration deviation ($\Delta$acc, rad/s$^2$) are used.
% During the recovery task, the robot is subjected to strong external perturbations that may cause unstable tracking, severe balance loss, or complete falls.
% Recovery is considered successful if the robot re-establishes stable tracking within the predefined tracking-error threshold after the disturbance or fall; otherwise, the trial is considered unsuccessful.
% The main evaluation metric is the success rate (Succ, \%).
\textbf{Evaluation Metrics and Training Details.}
% We evaluate motion tracking and fall recovery. 
For motion tracking, we use root-relative mean per-body position error (MPBPE, mm), mean per-joint position error (MPJPE, $10^{-3}$ rad), mean per-joint velocity error (MPJVE, $10^{-3}$ rad/s), and acceleration deviation ($\Delta$acc, rad/s$^2$). 
For recovery, the robot is subjected to strong external perturbations that may cause unstable tracking, severe balance loss, or complete falls. 
A trial is successful if the robot re-establishes stable tracking within the predefined tracking-error threshold after the disturbance or fall, and unsuccessful otherwise. 
The main evaluation metric is the success rate (Succ, \%). The reward items and key training configurations are summarized in the Appendix (see Tables~\ref{tab:reward_terms} and \ref{tab:domain_randomization}, respectively). 
\subsection{ Results on Motion Tracking }
To address \textbf{Q1}, this section compares the motion tracking accuracy of Stubborn with baseline methods on the full LAFAN1 dataset.
All methods were evaluated on the MuJoCo simulation platform, and the results are shown in Table~\ref{tab:main_results_motion_tracking}, where the results are reported as the mean $\pm$ one standard deviation.

Stubborn achieves lower errors in MPBPE, MPJPE, and MPJVE.
Compared with From-scratch RL, MPBPE decreases from $62.68$ to $48.85$, MPJPE decreases from $115.43$ to $113.38$, and MPJVE decreases from $887.87$ to $624.03$.
These results indicate that the proposed method improves whole-body position tracking and joint velocity tracking performance under the current test settings.
Furthermore, the relatively small standard deviations of MPBPE and MPJVE suggest that the evaluation results on the full LAFAN1 dataset are relatively stable.
%%%%%%%%%%%%%%%%%%%%%%%%%%%%%%%%%%%%%%%%%%%%%%%%%%%%%%%%%%%%%%%%%%%%%%%%%%%%%%%%%%%%%%%%%%%%%%%%%%%%%%%%%%%%%%%%%%%%%%
% TableX
\begin{table*}[h]
    \centering
    \caption{Quantitative results of motion tracking on the complete LAFAN1 dataset. Results are reported as mean $\pm$ standard deviation. Bold indicates the best result.}
    \label{tab:main_results_motion_tracking}  
    \resizebox{\textwidth}{!}{%
    \begin{tabular}{l*{6}{c}}
        \toprule
        \multicolumn{1}{c}{Method \textbar\ Metric} 
        & MPBPE $\downarrow$
        & MPJPE $\downarrow$ 
        & MPJVE $\downarrow$ 
        & $\Delta \text{acc} \downarrow$ \\
        \midrule
        HoloMotion~\cite{HoloMotion} 
        & $162.92_{\pm 0.000}$ 
        & $150.65{\pm 0.000}$ 
        & $758.56_{\pm 0.000}$
        & $19.47_{\pm0.000}$
        \\
        
        Any2Track~\cite{zhang2025track} 
        & $401.26_{\pm 0.000}$ 
        & $122.44_{\pm 0.000}$ 
        & $913.41_{\pm 0.000}$
        & $26.44_{\pm0.000}$
        \\
        
        BFM-Zero~\cite{li2025bfm}
        & $214.09_{\pm 0.000}$ 
        & $299.36_{\pm 0.000}$ 
        & $1100.88_{\pm 0.000}$
        & $\mathbf{17.07}_{\pm0.000}$
        \\
        
        From-scratch RL~\cite{liao2025beyondmimic}
        & $62.68_{\pm 0.000}$ 
        & $115.43_{\pm 0.000}$ 
        & $887.87_{\pm 0.000}$
        & $27.75_{\pm0.000}$
        \\
        
        Ours 
        & $\mathbf{48.85}_{\pm 0.000}$
        & $\mathbf{113.38}_{\pm 0.000}$ 
        & $\mathbf{624.03}_{\pm 0.000}$
        & $17.09_{\pm0.000}$
        \\
        \bottomrule
    \end{tabular}%
    }
\end{table*}
%%%%%%%%%%%%%%%%%%%%%%%%%%%%%%%%%%%%%%%%%%%%%%%%%%%%%%%%%%%%%%%%%%%%%%%%%%%%%%%%%%%%%%%%%%%%%%%%%%%%%%%%%%%%%%%%%%%%%%

For $\Delta \text{acc}$, Stubborn achieves a score of $17.09$, which is lower than HoloMotion, Any2Track, and From-scratch RL, and is close to BFM-Zero.
Although BFM-Zero achieves a slightly lower $\Delta \text{acc}$ value, its MPBPE, MPJPE, and MPJVE are substantially higher than those of Stubborn.
Considering these metrics together, Stubborn achieves lower tracking errors across the full LAFAN1 dataset while maintaining stable, smoother dynamic motion performance.

\subsection{ Ablation Studies }
% This section presents ablation studies to evaluate the contributions of the probabilistic termination mechanism and the tracking-error-driven adaptive sampling strategy in Stubborn.
%%%%%%%%%%%%%%%%%%%%%%%%%%%%%%%%%%%%%%%%%%%%%%%%%%%%%%%%%%%%%%%%%%%%%%%%%%%%%%%%
\begin{table}[b!]
    \centering
    \caption{\revised{Recovery success under 5 m/s perturbations. All methods are evaluated at iteration 90,000 with a 1,000-step recovery timeout. Bold indicates the highest success rate in each threshold column.}}
    \label{tab:ablation_pt}
    % \small
    % \setlength{\tabcolsep}{5pt}
    \renewcommand{\arraystretch}{1.0}
    \begin{tabular*}{0.95\columnwidth}{@{\extracolsep{\fill}}lcc@{}}
        \toprule
        \textbf{Method} & \multicolumn{2}{c}{\textbf{Success (\%)} $\uparrow$} \\
        \cmidrule(lr){2-3}
                       & \revised{0.15 m} & \revised{0.25 m} \\
        \midrule
        w/o PT  & \revised{55.0} & \revised{62.5} \\
        Ours    & \revised{90.0} & \textbf{100} \\
        \revised{Window}  & \revised{\textbf{97.5}} & \revised{97.5} \\
        \bottomrule
    \end{tabular*}
\end{table}
%%%%%%%%%%%%%%%%%%%%%%%%%%%%%%%%%%%%%%%%%%%%%%%%%%%%%%%%%%%%%%%%%%%%%%%%%%%%%%%%

\revised{To address \textbf{Q2}, we compare the full Stubborn framework (Ours), the variant without PT (w/o PT), and a fixed-window termination variant (Window) under 5 m/s perturbations.}
\revised{In Window, an episode terminates after $K=200$ consecutive training steps if at least one termination limit (the same as those in Eq. 2) is violated at each step within this consecutive window; returning within both training termination limits before the window deadline resets the failure counter and allows the episode to continue. This rule replaces Bernoulli termination with a deterministic recovery deadline.}
\revised{For continuous threshold violations, the default PT still has survival probability $P(\mathcal{T}>200)=0.995^{200}\approx36.7\%$, whereas Window terminates at its 200-step deadline.}
% Window K=200: accompanying DDT delivery README; this is a training counter,
% distinct from the evaluation threshold and evaluation timeout below.

\revised{We evaluate recovery at body tracking-error thresholds of 0.15 m and 0.25 m, with the former imposing a stricter criterion. Each checkpoint is tested with 40 perturbations, ten in each of four directions, on \texttt{dance1\_subject2}. Recovery must occur within 1,000 steps after the perturbation; mean recovery counts are computed only over successful trials.}
\revised{Table~\ref{tab:ablation_pt} reports recovery success at iteration 90,000, recomputed from the trial-level records. At 0.15 m, Ours, w/o PT, and Window achieve 90.0\%, 55.0\%, and 97.5\% success, respectively; at 0.25 m, they achieve 100\%, 62.5\%, and 97.5\%.}

%%%%%%%%%%%%%%%%%%%%%%%%%%%%%%%%%%%%%%%%%%%%%%%%%%%%%%%%%%%%%%%%%%%%%%%%%%%%%%%%%%%%%%%%%%%%%%%%%%%%%%%%%%%%%%%%%%%%%%%%%%%%%%%
%% Figure
\begin{figure*}[h!]
    \centering
    \includegraphics[width=0.99\textwidth]{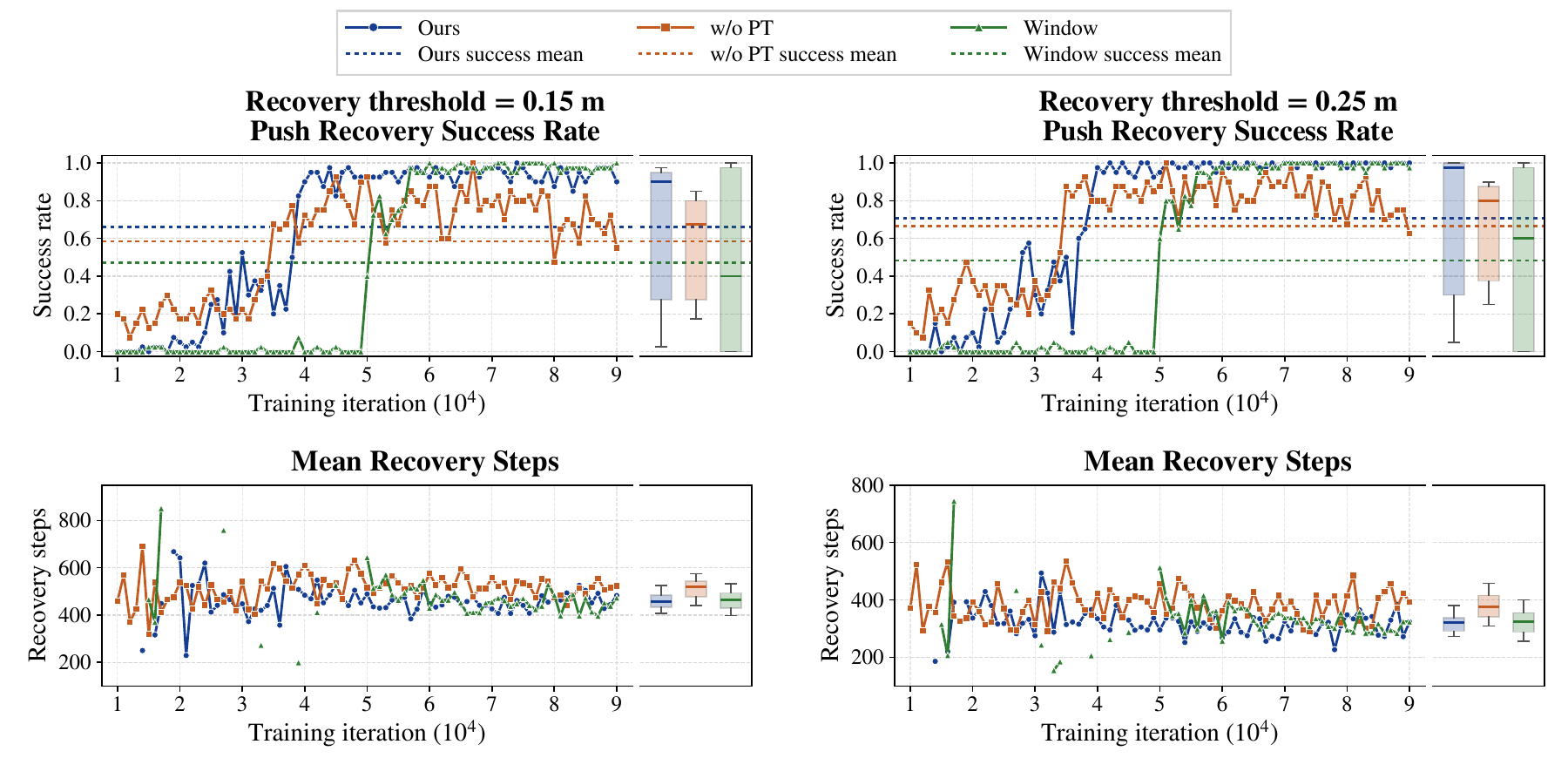}
    \caption{\revised{Recovery during training under 5 m/s perturbations for Ours, w/o PT, and Window. All curves use checkpoints from iterations 10,000 to 90,000 and a 1,000-step recovery cutoff. Recovery counts average successful trials only; zero-success checkpoints have no recovery-count value. At some early training checkpoints, Window fails to stand up and resume tracking in any trial (0\% success), leaving its mean recovery count undefined and causing gaps in the green Mean Recovery Steps curves. Dotted lines in the success-rate panels indicate checkpoint means. Box plots summarize the displayed finite checkpoint values: boxes show the interquartile range, internal lines the median, and whiskers the 10th and 90th percentiles. These distributions describe checkpoints within each run.}}
    \label{fig:ablation_push_recovery}
\end{figure*}
%%%%%%%%%%%%%%%%%%%%%%%%%%%%%%%%%%%%%%%%%%%%%%%%%%%%%%%%%%%%%%%%%%%%%%%

\revised{Figure~\ref{fig:ablation_push_recovery} compares the recovery performance throughout training. It can be seen that PT and Window improve final-checkpoint recovery over w/o PT for both body tracking-error thresholds of 0.15 m and 0.25 m. Among successful trials at iteration 90,000, Window's mean recovery counts are 472.6 and 323.4 steps at 0.15 m and 0.25 m, respectively.}

\revised{Qualitative rollouts of Window also suggest a reward-hacking tendency. The policy sometimes remains standing while the reference advances past a difficult segment, or skips the challenging motion and resumes tracking a subsequent reference segment. Because Window defers threshold-triggered termination until its deadline, the policy can avoid termination if the triggering errors return within their limits in time. Such behavior can allow difficult reference motions to be bypassed; a high recovery success rate therefore does not by itself demonstrate faithful tracking throughout the intervening segment.}

\revised{The default PT retains a termination risk at every threshold-violating step, rather than providing a fixed tolerant period as in the Window strategy. The runs are aligned by training iteration; matched cumulative environment-interaction budgets have not been established, so these trajectories alone do not establish relative sample efficiency.}

\textbf{Sensitivity to the termination probability parameter.}
We further evaluate the effect of the Bernoulli termination probability by training policies with
$p_{\mathrm{term}}\in\{0,0.005,0.01,0.05\}$ while keeping all other training configurations unchanged.
Here, $p_{\mathrm{term}}=0.005$ is the default setting of Stubborn.
Setting $p_{\mathrm{term}}=0$ disables threshold-triggered probabilistic termination, while retaining the maximum episode-length limit.
All variants are evaluated using their checkpoints at training iteration 52,000.

As shown in Table~\ref{tab:ablation_pterm_sensitivity}, the variants with $p_{\mathrm{term}}=0.01$ and $p_{\mathrm{term}}
=0.05$ achieve tracking accuracy close to the default setting, suggesting limited sensitivity over the tested nonzero-probability range.
The default setting of $p_{\mathrm{term}}=0.005$ achieves the lowest MPBPE of 48.91 mm, whereas removing probabilistic termination ($p_{\mathrm{term}}=0$) increases MPBPE to
56.26 mm.

%%%%%%%%%%%%%%%%%%%%%%%%%%%%%%%%%%%%%%%%%%%%%%%%%%%%%%%%%%%%%%%%%%%%%%%%%%%%%%%%
\begin{table}[t!]
\centering
\caption{Sensitivity analysis of the termination probability
$p_{\mathrm{term}}$. All variants are evaluated at training iteration
52,000. The default setting is $p_{\mathrm{term}}=0.005$.
Bold indicates the lowest error.}
\label{tab:ablation_pterm_sensitivity}
\renewcommand{\arraystretch}{1.0}
\resizebox{0.95\columnwidth}{!}{%
\begin{tabular}{ccccc}
    \toprule
    $p_{\mathrm{term}}$ & MPBPE $\downarrow$ & MPJPE $\downarrow$ & MPJVE $\downarrow$ & $\Delta\mathrm{acc}\downarrow$ \\
    \midrule
    $0$     & 56.26 & \textbf{98.5} & 637.6 & 17.46 \\
    $0.005$ & \textbf{48.91} & 117.4 & 634.9 & 17.21 \\
    $0.01$  & 50.73 & 115.0 & \textbf{632.0} & \textbf{17.08} \\
    $0.05$  & 52.48 & 116.6 & 634.3 & 17.21 \\
    \bottomrule
\end{tabular}}
\end{table}

%%%%%%%%%%%%%%%%%%%%%%%%%%%%%%%%%%%%%%%%%%%%%%%%%%%%%%%%%%%%%%%%%%%%%%%%%%%%%%%%
\begin{figure}[t!]
\centering
\includegraphics[width=0.99\columnwidth]{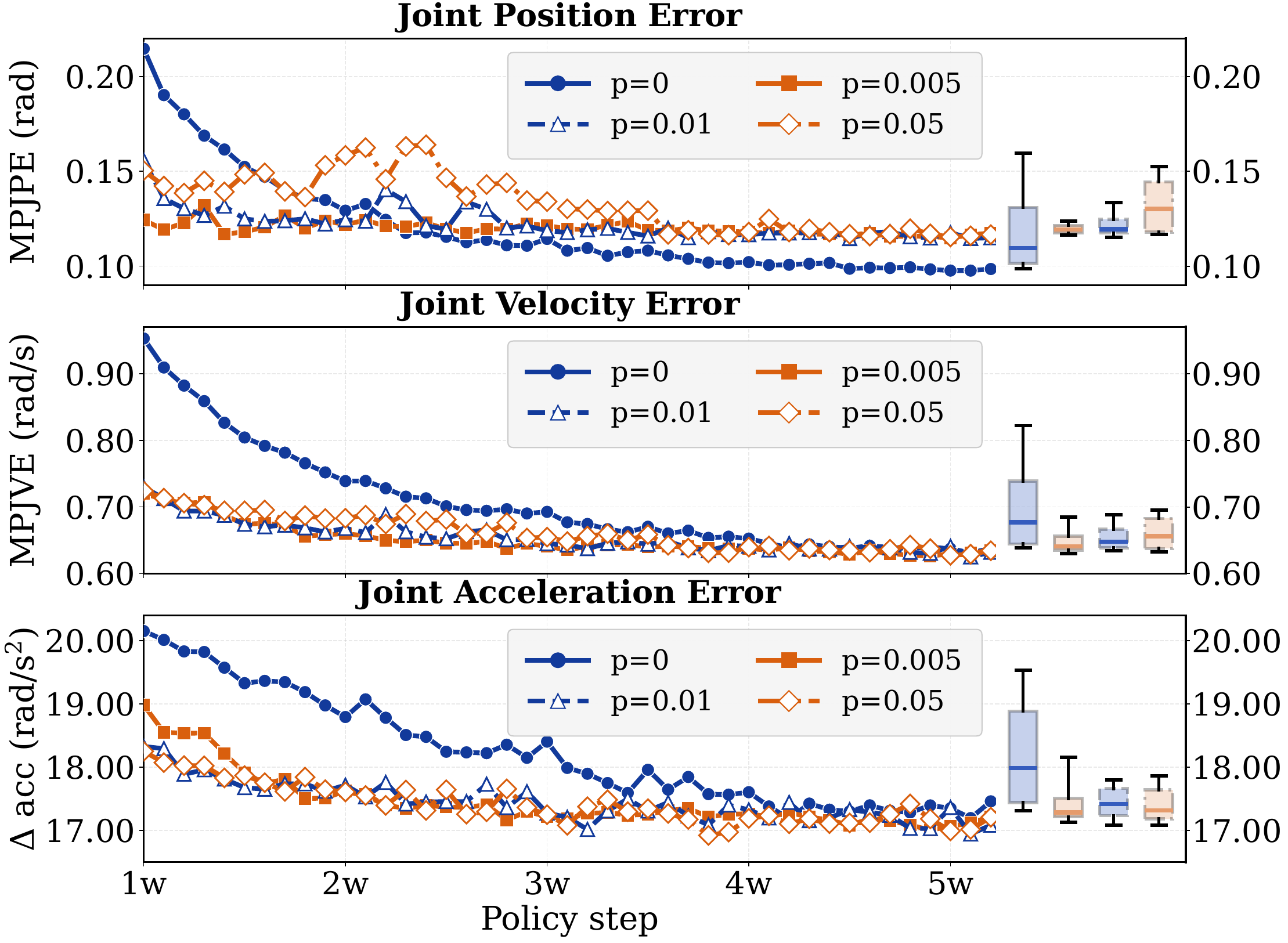}
{
\caption{Tracking errors during training for different termination probabilities.
All curves show checkpoints from iterations 10,000 to 52,000.
The box plots summarize the displayed checkpoints: boxes indicate the
interquartile range, the internal line denotes the median, and whiskers
correspond to the 10th and 90th percentiles.}
\label{fig:ablation_pterm_tracking}
}
\end{figure}
%%%%%%%%%%%%%%%%%%%%%%%%%%%%%%%%%%%%%%%%%%%%%%%%%%%%%%%%%%%%%%%%%%%%%%%%%%%%%%%%

Figure~\ref{fig:ablation_pterm_tracking} further compares tracking performance during training.
The default setting exhibits lower joint-position errors early in the displayed training interval and the smallest
interquartile ranges across all three plotted metrics, indicating a more concentrated error distribution across the
evaluated checkpoints.

In comparison, $p_{\mathrm{term}}=0$ exhibits larger errors early in training.
Qualitative rollout observations further suggest that, when $p_{\mathrm{term}}=0$, the policy may favor simpler, more
easily stabilized motions over difficult reference segments, thereby leading to the lowest MPJPE (see Table~\ref{tab:ablation_pterm_sensitivity}). This behavior is consistent with a reward-hacking tendency, in which maintaining a stable alternative motion takes
precedence over faithfully following the reference trajectory.
% A nonzero termination probability provides a mechanism for limiting prolonged threshold violations while preserving an opportunity for recovery.
Together with the recovery-time interpretation in Section~\ref{subsec:probabilistic_termination}, these observations support
the choice of $p_{\mathrm{term}}=0.005$ as a practical balance between recovery-oriented exploration and whole-body tracking
accuracy.
%%%%%%%%%%%%%%%%%%%%%%%%%%%%%%%%%%%%%%%%%%%%%%%%%%%%%%%%%%%%%%%%%%%%%%%%%%%%%%%%
\begin{table}[h!]
    \centering
    \caption{Ablation study of our proposed sampling strategy. Bold means better.}
    \label{tab:ablation_adps}
    \renewcommand{\arraystretch}{1.0}
    \resizebox{0.80\columnwidth}{!}{
    \begin{tabular}{cccc}
        \toprule
        \textbf{Method} & \textbf{MPJPE} $\downarrow$ & \textbf{MPJVE} $\downarrow$  & $\Delta \textbf{acc} \downarrow$ \\
        \midrule
        w/o AdpS & 120.7 & 548.5 & 14.50 \\
        Ours     & \textbf{119.4} & \textbf{541.5} & \textbf{14.42}   \\
        \bottomrule
    \end{tabular}
    }
\end{table}

% \vspace{-0.4cm} % removed for ICRA formatting compliance
%%%%%%%%%%%%%%%%%%%%%%%%%%%%%%%%%%%%%%%%%%%%%%%%%%%%%%%%%%%%%%%%%%%%%%%%%%%%%%%%

To address \textbf{Q3}, we perform an ablation study on the effect of the adaptive PT and tracking error-driven sampling strategy (AdpS) by comparing the full Stubborn with a variant without this module (w/o AdpS), while keeping all other settings unchanged.
As shown in Table~\ref{tab:ablation_adps}, introducing AdpS reduces MPJPE, MPJVE, and $\Delta {\mathrm{acc}}$ from 120.7, 548.5, and 14.50 to 119.4, 541.5, and 14.42, respectively, indicating improved tracking performance in terms of joint position, joint velocity, and acceleration.

\begin{figure}[t!]
    \centering
    \includegraphics[width=0.99\columnwidth]{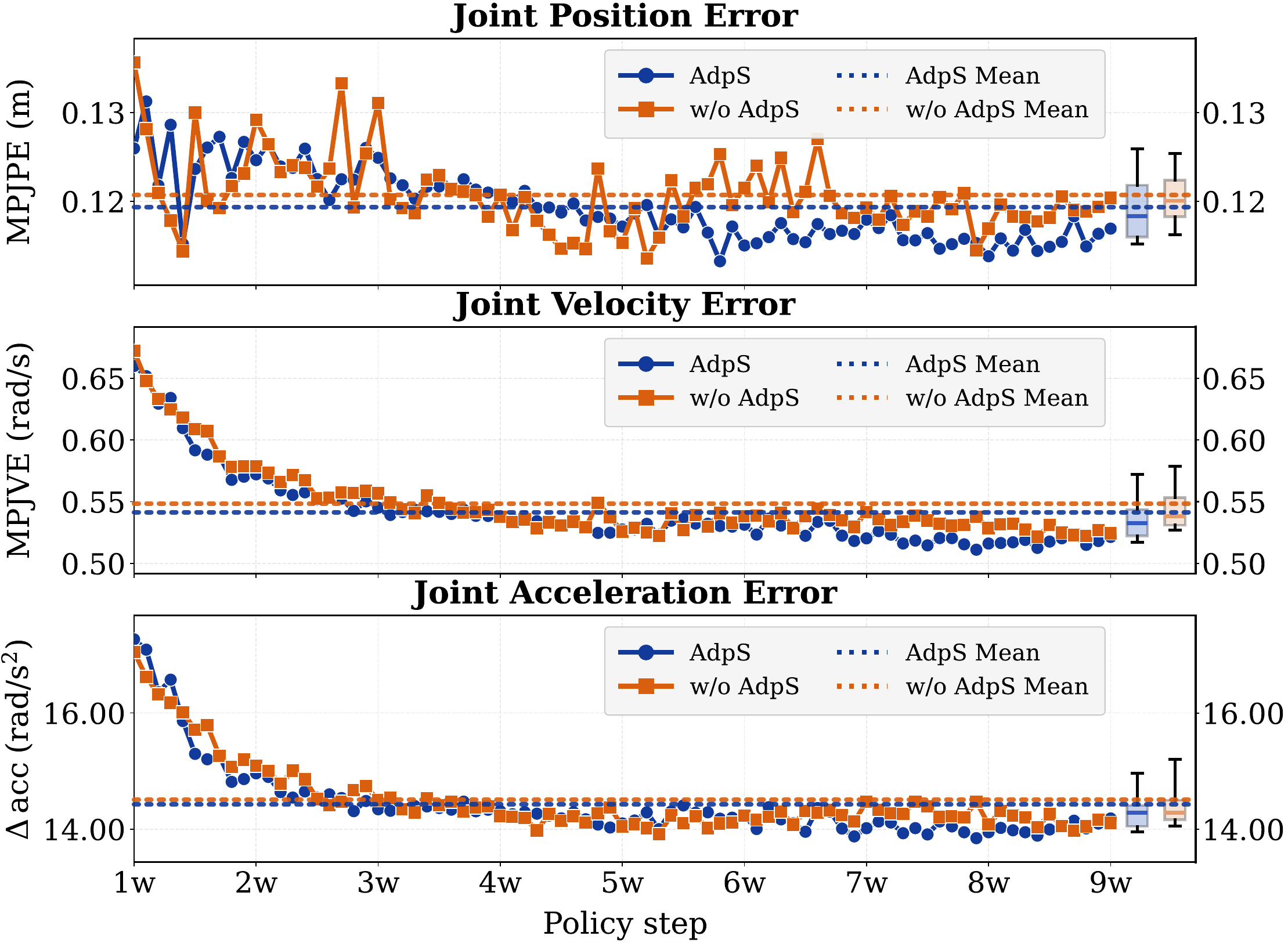}
    \caption{ Ablation results of the AdpS sampling strategy. 
    % The figure shows the training evolution of MPJPE, MPJVE, and $\Delta \text{acc}$ with and without AdpS. 
    The results suggest that AdpS improves motion-tracking performance in the later training stages.}
    \label{fig:ablation_AdpS}
\end{figure}

Fig.~\ref{fig:ablation_AdpS} further illustrates the error trends during training.
AdpS maintains lower joint position and velocity errors overall in the middle and late stages, while the acceleration error remains similar to that of w/o AdpS. 
The box plot on the right summarizes the error distribution in the final training stage, with the solid line within each box indicating the median. 
These results suggest that tracking-error-based adaptive sampling improves the overall tracking performance of long-horizon, highly dynamic motions by increasing the training frequency for difficult segments.

%%%%%%%%%%%%%%%%%%%%%%%%%%%%%%%%%%%%%%%%%%%%%%%%%%%%%%%%%%%%%%%%%%%%%%%%%%%%%%%%%%%%%%%%%%%%%%%%%%%%%%%%%%%%%%%%%%%%%%%%%%%%%%%
%% Figure
%%%%%%%%%%%%%%%%%%%%%%%%%%%%%%%%%%%%%%%%%%%%%%%%%%%%%%%%%%%%%%%%%%%%%%%%%%%%%%%%%%%%%%%%%%%%%%%%%%%%%%%%%%%%%%%%%%%%%%%%%%%%%%%

\subsection{Real-World Deployment}
\begin{figure}[h!]
    \centering
    \includegraphics[width=0.99\columnwidth]{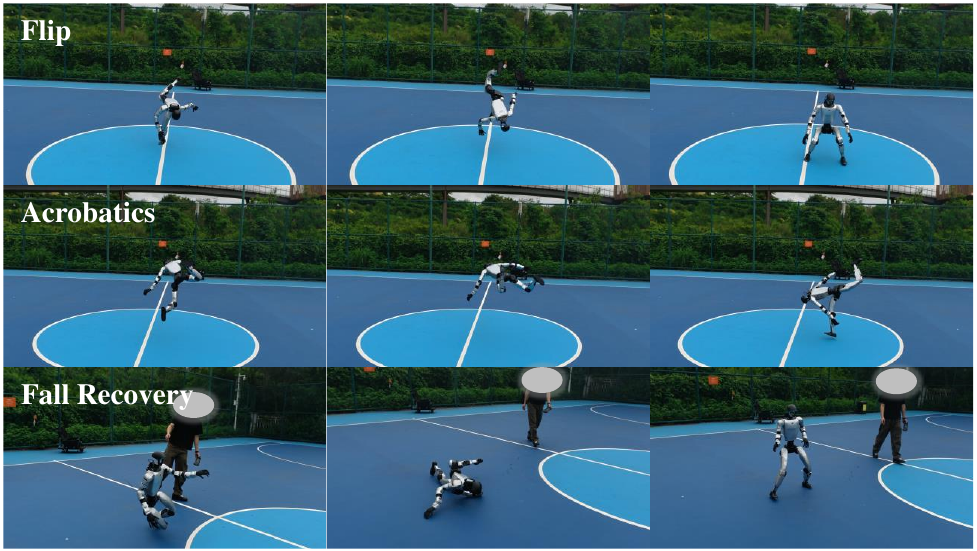}
    \caption{Experimental results. Stubborn can recover from falls and precisely tracks diverse dynamic motions, including flips and acrobatic maneuvers. 
    % Physical experiments further demonstrate the learned policy's robustness and fall recovery against external disturbances.
    }
    \label{fig:real_world_g1}
\end{figure}

% To address \textbf{Q4}, we evaluate the real-world deployment viability of Stubborn on the physical Unitree G1 humanoid platform (as shown in Figure~\ref{fig:real_world_g1}).
% The hardware experiments demonstrate the robot's locomotion stability under external disturbances, alongside its fall recovery performance. The results indicate that Stubborn maintains reliable motion tracking performance and robustness in physical environments.
To address \textbf{Q4}, we evaluate the real-world deployment viability of Stubborn on the physical 29-DoF Unitree G1 humanoid platform, as shown in Fig.~\ref{fig:real_world_g1}. 
The real-robot experiments include motions from LAFAN1~\cite{harvey2020robust} and AMASS~\cite{mahmood2019amass}, and demonstrate the robot's motion tracking, disturbance robustness, and fall recovery performance in physical environments.
% Additional video demonstrations of the hardware deployments are available on the project website.
%%%%%%%%%%%%%%%%%%%%%%%%%%%%%%%%%%%%%%%%%%%%%%%%%%%%%%%%%%%%%%%%%%%%%%%%%%%%%%%%%%%%%%%%%%%%%%%%%%%%%%%%%%%%%%%%%%%%%%%%%%%%%%%
%% Figure
%%%%%%%%%%%%%%%%%%%%%%%%%%%%%%%%%%%%%%%%%%%%%%%%%%%%%%%%%%%%%%%%%%%%%%%%%%%%%%%%%%%%%%%%%%%%%%%%%%%%%%%%%%%%%%%%%%%%%%%%%%%%%%%

\section{Conclusion}
\label{sec:conclusion}
This paper presents Stubborn, a streamlined
and unified RL framework that enables the learning of one policy for robust motion tracking and fall recovery for humanoids.
By adopting a yaw-aligned tracking representation, we have proposed a PT mechanism, together with a PT and tracking error-driven sampling strategy. The proposed method improves tracking robustness and enables recovery from both unstable and fallen states under strong perturbations.
Experimental results in simulation and on the Unitree G1 humanoid robot demonstrate that Stubborn offers effective disturbance recovery, fall recovery, and competitive motion tracking performance under challenging conditions, in comparison with STOA methods. Future work will focus on improving the agility and smoothness of recovery motions after severe falls, as well as extending the framework to more challenging terrains and highly dynamic humanoid behaviors.

\section*{\revised{Acknowledgment}}
\revised{We thank Shaochen Mei for his assistance in the fall-recovery experiments associated with the fixed-window termination strategy.}

%===============================================================================
% References
\bibliographystyle{IEEEtran}
\bibliography{example}

@article{peng2018deepmimic,
  title={Deepmimic: Example-guided deep reinforcement learning of physics-based character skills},
  author={Peng, Xue Bin and Abbeel, Pieter and Levine, Sergey and Van de Panne, Michiel},
  journal={ACM Transactions On Graphics (TOG)},
  volume={37},
  number={4},
  pages={1--14},
  year={2018},
  publisher={ACM New York, NY, USA}
}

@article{liao2025beyondmimic,
  title={Beyondmimic: From motion tracking to versatile humanoid control via guided diffusion},
  author={Liao, Qiayuan and Truong, Takara E and Huang, Xiaoyu and Gao, Yuman and Tevet, Guy and Sreenath, Koushil and Liu, C Karen},
  journal={arXiv preprint arXiv:2508.08241},
  year={2025}
}

@article{fu2024humanplus,
  title={Humanplus: Humanoid shadowing and imitation from humans},
  author={Fu, Zipeng and Zhao, Qingqing and Wu, Qi and Wetzstein, Gordon and Finn, Chelsea},
  journal={arXiv preprint arXiv:2406.10454},
  year={2024}
}

@article{he2024omnih2o,
  title={Omnih2o: Universal and dexterous human-to-humanoid whole-body teleoperation and learning},
  author={He, Tairan and Luo, Zhengyi and He, Xialin and Xiao, Wenli and Zhang, Chong and Zhang, Weinan and Kitani, Kris and Liu, Changliu and Shi, Guanya},
  journal={arXiv preprint arXiv:2406.08858},
  year={2024}
}

@article{he2025asap,
  title={Asap: Aligning simulation and real-world physics for learning agile humanoid whole-body skills},
  author={He, Tairan and Gao, Jiawei and Xiao, Wenli and Zhang, Yuanhang and Wang, Zi and Wang, Jiashun and Luo, Zhengyi and He, Guanqi and Sobanbab, Nikhil and Pan, Chaoyi and others},
  journal={arXiv preprint arXiv:2502.01143},
  year={2025}
}

@article{xu2026reach,
  title={REACH: An RL-based Hierarchical Humanoid Crawling Framework for Adaptable Confined Space Traversal},
  author={Xu, Shengzhang and Guo, Yuning and Hu, Weiyi and Xu, Yifeng and Li, Zhongyu and Kong, He},
  journal={arXiv preprint arXiv:2610.36546},
  year={2026}
}

@article{chen2025gmt,
  title={Gmt: General motion tracking for humanoid whole-body control},
  author={Chen, Zixuan and Ji, Mazeyu and Cheng, Xuxin and Peng, Xuanbin and Peng, Xue Bin and Wang, Xiaolong},
  journal={arXiv preprint arXiv:2506.14770},
  year={2025}
}

@article{he2025learning,
  title={Learning getting-up policies for real-world humanoid robots},
  author={He, Xialin and Dong, Runpei and Chen, Zixuan and Gupta, Saurabh},
  journal={arXiv preprint arXiv:2502.12152},
  year={2025}
}

@article{huang2502learning,
  title={Learning humanoid standing-up control across diverse postures (2025)},
  author={Huang, T and Ren, J and Wang, H and Wang, Z and Ben, Q and Wen, M and Chen, X and Li, J and Pang, J},
  journal={URL https://arxiv. org/abs/2502.08378}
}

@article{cheng2024expressive,
  title={Expressive whole-body control for humanoid robots},
  author={Cheng, Xuxin and Ji, Yandong and Chen, Junming and Yang, Ruihan and Yang, Ge and Wang, Xiaolong},
  journal={arXiv preprint arXiv:2402.16796},
  year={2024}
}

@article{ji2024exbody2,
  title={Exbody2: Advanced expressive humanoid whole-body control},
  author={Ji, Mazeyu and Peng, Xuanbin and Liu, Fangchen and Li, Jialong and Yang, Ge and Cheng, Xuxin and Wang, Xiaolong},
  journal={arXiv preprint arXiv:2412.13196},
  year={2024}
}

@article{luo2025sonic,
  title={Sonic: Supersizing motion tracking for natural humanoid whole-body control},
  author={Luo, Zhengyi and Yuan, Ye and Wang, Tingwu and Li, Chenran and Chen, Sirui and Castaneda, Fernando and Cao, Zi-Ang and Li, Jiefeng and Minor, David and Ben, Qingwei and others},
  journal={arXiv preprint arXiv:2511.07820},
  year={2025}
}

@article{yin2025unitracker,
  title={Unitracker: Learning universal whole-body motion tracker for humanoid robots},
  author={Yin, Kangning and Zeng, Weishuai and Fan, Ke and Dai, Minyue and Wang, Zirui and Zhang, Qiang and Tian, Zheng and Wang, Jingbo and Pang, Jiangmiao and Zhang, Weinan},
  journal={arXiv preprint arXiv:2507.07356},
  year={2025}
}

@article{liu2025opt2skill,
  title={Opt2skill: Imitating dynamically-feasible whole-body trajectories for versatile humanoid loco-manipulation},
  author={Liu, Fukang and Gu, Zhaoyuan and Cai, Yilin and Zhou, Ziyi and Jung, Hyunyoung and Jang, Jaehwi and Zhao, Shijie and Ha, Sehoon and Chen, Yue and Xu, Danfei and others},
  journal={IEEE Robotics and Automation Letters},
  year={2025},
  publisher={IEEE}
}

@article{radosavovic2024real,
  title={Real-world humanoid locomotion with reinforcement learning},
  author={Radosavovic, Ilija and Xiao, Tete and Zhang, Bike and Darrell, Trevor and Malik, Jitendra and Sreenath, Koushil},
  journal={Science Robotics},
  volume={9},
  number={89},
  pages={eadi9579},
  year={2024},
  publisher={American Association for the Advancement of Science}
}

@article{wang2026omnixtreme,
  title={OmniXtreme: Breaking the Generality Barrier in High-Dynamic Humanoid Control},
  author={Wang, Yunshen and Zhu, Shaohang and Zhi, Peiyuan and Li, Yuhan and Li, Jiaxin and Li, Yong-Lu and Xiao, Yuchen and Wang, Xingxing and Jia, Baoxiong and Huang, Siyuan},
  journal={arXiv preprint arXiv:2602.23843},
  year={2026}
}

@inproceedings{li2025clone,
  title={Clone: Closed-loop whole-body humanoid teleoperation for long-horizon tasks},
  author={Li, Yixuan and Lin, Yutang and Cui, Jieming and Liu, Tengyu and Liang, Wei and Zhu, Yixin and Huang, Siyuan},
  booktitle={9th Annual Conference on Robot Learning},
  year={2025}
}

@article{zhang2025hub,
  title={Hub: Learning extreme humanoid balance},
  author={Zhang, Tong and Zheng, Boyuan and Nai, Ruiqian and Hu, Yingdong and Wang, Yen-Jen and Chen, Geng and Lin, Fanqi and Li, Jiongye and Hong, Chuye and Sreenath, Koushil and others},
  journal={arXiv preprint arXiv:2505.07294},
  year={2025}
}

@article{wu2025learn,
  title={Learn to teach: Sample-efficient privileged learning for humanoid locomotion over real-world uneven terrain},
  author={Wu, Feiyang and Nal, Xavier and Jang, Jaehwi and Zhu, Wei and Gu, Zhaoyuan and Wu, Anqi and Zhao, Ye},
  journal={IEEE Robotics and Automation Letters},
  year={2025},
  publisher={IEEE}
}

@article{lee2025learning,
  title={Learning Humanoid Arm Motion via Centroidal Momentum Regularized Multi-Agent Reinforcement Learning},
  author={Lee, Ho Jae and Jeon, Se Hwan and Kim, Sangbae},
  journal={IEEE Robotics and Automation Letters},
  year={2025},
  publisher={IEEE}
}

@article{jung2025ppf,
  title={PPF: Pre-training and Preservative Fine-tuning of Humanoid Locomotion via Model-Assumption-based Regularization},
  author={Jung, Hyunyoung and Gu, Zhaoyuan and Zhao, Ye and Park, Hae-Won and Ha, Sehoon},
  journal={IEEE Robotics and Automation Letters},
  year={2025},
  publisher={IEEE}
}

@article{cheng2025rambo,
  title={RAMBO: RL-augmented Model-based Whole-body Control for Loco-manipulation},
  author={Cheng, Jin and Kang, Dongho and Fadini, Gabriele and Shi, Guanya and Coros, Stelian},
  journal={IEEE Robotics and Automation Letters},
  year={2025},
  publisher={IEEE}
}

@article{han2025kungfubot2,
  title={cc: Learning versatile motion skills for humanoid whole-body control},
  author={Han, Jinrui and Xie, Weiji and Zheng, Jiakun and Shi, Jiyuan and Zhang, Weinan and Xiao, Ting and Bai, Chenjia},
  journal={arXiv preprint arXiv:2509.16638},
  year={2025}
}

@article{xie2025kungfubot,
  title={Kungfubot: Physics-based humanoid whole-body control for learning highly-dynamic skills},
  author={Xie, Weiji and Han, Jinrui and Zheng, Jiakun and Li, Huanyu and Liu, Xinzhe and Shi, Jiyuan and Zhang, Weinan and Bai, Chenjia and Li, Xuelong},
  journal={arXiv preprint arXiv:2506.12851},
  year={2025}
}

@article{peng2021amp,
  title={Amp: Adversarial motion priors for stylized physics-based character control},
  author={Peng, Xue Bin and Ma, Ze and Abbeel, Pieter and Levine, Sergey and Kanazawa, Angjoo},
  journal={ACM Transactions on Graphics (ToG)},
  volume={40},
  number={4},
  pages={1--20},
  year={2021},
  publisher={ACM New York, NY, USA}
}

@inproceedings{luo2023perpetual,
  title={Perpetual Humanoid Control for Real-time Simulated Avatars},
  author={Luo, Zhengyi and Cao, Jinkun and Winkler, Alexander and Kitani, Kris and Xu, Weipeng},
  booktitle={Proceedings of the IEEE/CVF International Conference on Computer Vision},
  pages={10895--10904},
  year={2023}
}

@article{sun2025robotdancing,
  title={RobotDancing: Residual-Action Reinforcement Learning Enables Robust Long-Horizon Humanoid Motion Tracking},
  author={Sun, Zhenguo and Peng, Yibo and Meng, Yuan and Li, Xukun and Huang, Bo-Sheng and Bing, Zhenshan and Wang, Xinlong and Knoll, Alois},
  journal={arXiv preprint arXiv:2509.20717},
  year={2025}
}

@inproceedings{mahmood2019amass,
  title={AMASS: Archive of motion capture as surface shapes},
  author={Mahmood, Naureen and Ghorbani, Nima and Troje, Nikolaus F and Pons-Moll, Gerard and Black, Michael J},
  booktitle={Proceedings of the IEEE/CVF international conference on computer vision},
  pages={5442--5451},
  year={2019}
}

@article{zhang2025track,
  title={Track any motions under any disturbances},
  author={Zhang, Zhikai and Guo, Jun and Chen, Chao and Wang, Jilong and Lin, Chenghuai and Lian, Yunrui and Xue, Han and Wang, Zhenrong and Liu, Maoqi and Lyu, Jiangran and others},
  journal={arXiv preprint arXiv:2509.13833},
  year={2025}
}

@article{ze2025twist,
  title={Twist: Teleoperated whole-body imitation system},
  author={Ze, Yanjie and Chen, Zixuan and Ara{\'u}jo, Joao Pedro and Cao, Zi-ang and Peng, Xue Bin and Wu, Jiajun and Liu, C Karen},
  journal={arXiv preprint arXiv:2505.02833},
  year={2025}
}

@article{ze2025twist2,
  title={Twist2: Scalable, portable, and holistic humanoid data collection system},
  author={Ze, Yanjie and Zhao, Siheng and Wang, Weizhuo and Kanazawa, Angjoo and Duan, Rocky and Abbeel, Pieter and Shi, Guanya and Wu, Jiajun and Liu, C Karen},
  journal={arXiv preprint arXiv:2511.02832},
  year={2025}
}

@inproceedings{chen2025hifar,
  title={Hifar: Multi-stage curriculum learning for high-dynamics humanoid fall recovery},
  author={Chen, Penghui and Wang, Yushi and Luo, Changsheng and Cai, Wenhan and Zhao, Mingguo},
  booktitle={2025 IEEE/RSJ International Conference on Intelligent Robots and Systems (IROS)},
  pages={2908--2915},
  year={2025},
  organization={IEEE}
}

@inproceedings{gaspard2025frasa,
  title={Frasa: An end-to-end reinforcement learning agent for fall recovery and stand up of humanoid robots},
  author={Gaspard, Cl{\'e}ment and Duclusaud, Marc and Passault, Gr{\'e}goire and Daniel, M{\'e}lodie and Ly, Olivier},
  booktitle={2025 IEEE International Conference on Robotics and Automation (ICRA)},
  pages={15994--16000},
  year={2025},
  organization={IEEE}
}

@inproceedings{egle2024enhancing,
  title={Enhancing model-based step adaptation for push recovery through reinforcement learning of step timing and region},
  author={Egle, Tobias and Yan, Yashuai and Lee, Dongheui and Ott, Christian},
  booktitle={2024 IEEE-RAS 23rd International Conference on Humanoid Robots (Humanoids)},
  pages={165--172},
  year={2024},
  organization={IEEE}
}

@inproceedings{duburcq2022reactive,
  title={Reactive stepping for humanoid robots using reinforcement learning: Application to standing push recovery on the exoskeleton atalante},
  author={Duburcq, Alexis and Schramm, Fabian and Bo{\'e}ris, Guilhem and Bredeche, Nicolas and Chevaleyre, Yann},
  booktitle={2022 IEEE/RSJ international conference on intelligent robots and systems (IROS)},
  pages={9302--9309},
  year={2022},
  organization={IEEE}
}

@inproceedings{yang2025bracing,
  title={Bracing for Impact: Robust Humanoid Push Recovery and Locomotion with Reduced Order Models},
  author={Yang, Lizhi and Werner, Blake and Ghansah, Adrian B and Ames, Aaron D},
  booktitle={2025 IEEE-RAS 24th International Conference on Humanoid Robots (Humanoids)},
  pages={728--735},
  year={2025},
  organization={IEEE}
}

@article{harvey2020robust,
  title={Robust motion in-betweening},
  author={Harvey, F{\'e}lix G and Yurick, Mike and Nowrouzezahrai, Derek and Pal, Christopher},
  journal={ACM Transactions on Graphics (TOG)},
  volume={39},
  number={4},
  pages={60--1},
  year={2020},
  publisher={ACM New York, NY, USA}
}

@article{li2025bfm,
  title={Bfm-zero: A promptable behavioral foundation model for humanoid control using unsupervised reinforcement learning},
  author={Li, Yitang and Luo, Zhengyi and Zhang, Tonghe and Dai, Cunxi and Kanervisto, Anssi and Tirinzoni, Andrea and Weng, Haoyang and Kitani, Kris and Guzek, Mateusz and Touati, Ahmed and others},
  journal={arXiv preprint arXiv:2511.04131},
  year={2025}
}

@software{HoloMotion,
  author = {Maiyue Chen and Kaihui Wang and Bo Zhang and Yi Ren and Zihao Zhu and Xihan Ma and Qijun Huang and Zhiyuan Yang and Yucheng Wang and Zhizhong Su},
  title = {HoloMotion: A Foundation Model for Whole-Body Humanoid Control},
  year = {2026},
  month = April,
  version = {1.2.0},
  url = {https://github.com/HorizonRobotics/HoloMotion},
  license = {Apache-2.0}
}

@article{zhu2026clot,
  title={CLOT: Closed-Loop Global Motion Tracking for Whole-Body Humanoid Teleoperation},
  author={Zhu, T. and Cai, G. and Yang, Z. and Ren, G. and Xie, H. and Wang, Z. and Wu, J. and Wang, J. and Yang, X. and Mu, Y. and Yan, Y.},
  journal={arXiv preprint arXiv:2602.15060},
  year={2026}
}

%===============================================================================

\section{Appendix}
\subsection{Drift-Invariant Yaw-Aligned Tracking Representation}
% Directly optimizing joint-space or global task space tracking errors often results in distorted postures due to human-robot kinematic heterogeneity and the required torques for disturbance-induced global translation and yaw deviation corrections.
% ; tracking global absolute coordinates requires torque to correct global translation and yaw deviations when the policy is perturbed, leading to system imbalance.
To decouple global drift from the desired motion style, similar to~\cite{cheng2025rambo, lee2025learning}, our framework adopts a yaw-invariant tracking representation based on a yaw-angle-aligned root coordinate system. 
% For completeness, the essential details of the yaw-invariant tracking representation are provided below. 
Let $W$ denote the world frame and $B$ the robot base frame.
% The global pose of the base $T_B^W \in SE(3)$ consists of the rotation $R_B^W \in SO(3)$ and the position $p_B^W \in \mathbb{R}^3$.
The global pose of the robot base remains defined in the world frame, where $T_B^W \in SE(3)$ consists of the rotation $R_B^W \in SO(3)$ and the position $p_B^W \in \mathbb{R}^3$.
Using the Z-Y-X Euler-angle convention, $R_B^W$ is decomposed into the product of the yaw angle $\psi$, pitch angle $\theta$, and roll angle $\phi$ as $R_B^W = R_Z(\psi) R_Y(\theta) R_X(\phi).$
% \begin{equation}
%     R_B^W = R_Z(\psi) R_Y(\theta) R_X(\phi).
% \end{equation}
Accordingly, we define the yaw-aligned frame $\Psi$, whose homogeneous transformation with respect to $W$ is given by 
$T_\Psi^W = \begin{bmatrix} R_Z(\psi) & p_B^W \\ 0_{1 \times 3} & 1 \end{bmatrix}.$
% \begin{equation*}
%     T_\Psi^W = \begin{bmatrix} R_Z(\psi) & p_B^W \\ 0_{1 \times 3} & 1 \end{bmatrix}.
% \end{equation*}
For any rigid body $i \in \mathcal{B}$, with world-frame pose $(R_i^W, p_i^W)$, its pose in $\Psi$ is obtained via 
\begin{equation*}
    T_i^\Psi = (T_\Psi^W)^{-1} T_i^W = \begin{bmatrix} R_Z^T(\psi) R_i^W & R_Z^T(\psi) (p_i^W - p_B^W) \\ 0_{1 \times 3} & 1 \end{bmatrix} .
\end{equation*}

The corresponding local position $\tilde{p}_{i}$ and orientation $\tilde{R}_{i}$ in the aligned frame are then $\tilde{p}_{i} = R_Z^T(\psi) (p_i^W - p_B^W), \quad \tilde{R}_{i} = R_Z^T(\psi) R_i^W.$
% \begin{equation}
%     \tilde{p}_{i} = R_Z^T(\psi) (p_i^W - p_B^W), \quad \tilde{R}_{i} = R_Z^T(\psi) R_i^W .
% \end{equation}
This representation reduces sensitivity to global drift while preserving key physical cues for motion control.
% The relative displacement term $(p_i^W - p_B^W)$ removes global translation, while the inverse yaw rotation $R_Z^\top(\psi)$ decouples global heading variation from local body coordination. Therefore, the resulting tracking errors depend primarily on local body coordination rather than global drift, and the policy is less likely to expend excessive control effort on unnecessary global correction under strong perturbations.
The relative displacement term $(p_i^W - p_B^W)$ removes global translation, while the inverse yaw rotation $R_Z^\top(\psi)$ removes the global heading component from the body pose representation. 
Therefore, the resulting tracking errors mainly reflect local whole-body coordination rather than global translation or yaw drift, reducing unnecessary global correction under strong perturbations. 
The yaw-aligned transformation is applied to the body tracking representation, while the root pose remains defined in the world frame, enabling the policy to reduce sensitivity to global drift while preserving gravity-related pitch and roll information.

% Moreover, the yaw-aligned tracking representation retains gravity-related orientation information. 
% Specifically, the base orientation in the $\Psi$ frame becomes $    \tilde{R}_{B} = R_Y(\theta) R_X(\phi)$,
% % \begin{equation}
% %     \tilde{R}_{B} = R_Y(\theta) R_X(\phi), 
% % \end{equation}
% which preserves pitch and roll while keeping the vertical axis consistent with the gravity direction. This enables the policy to remain sensitive to center-of-mass shifts and postural imbalance, thereby providing useful physical priors for subsequent fall-recovery learning.
% (see Section~\ref{subsec:probabilistic_termination}).

\subsection{Training Details of Stubborn}

% \vspace{-5cm} % removed for ICRA formatting compliance

% \subsection{Training Details of Stubborn}

%%%%%%%%%%%%%%%%%%%%%%%%%%%%%%%%%%%%%%%%%%
\begin{table}[h!]
\centering

\begin{minipage}{0.48\linewidth}
\centering
\caption{Reward terms for motion tracking.}
\label{tab:reward_terms}

\resizebox{\linewidth}{!}{
\begin{tabular}{llc}
\toprule
\textbf{Term} & \textbf{Formulation} & \textbf{Weight} \\
\midrule
Joint Position & $\exp\!\left(-\|p_{\mathrm{j}}\|^{2}/0.1\right)$ & 1.00 \\
Joint Velocity & $\exp\!\left(-\|v_{\mathrm{j}}\|^{2}/5\right)$ & 0.50 \\
Body Linear Velocity & $\exp\!\left(-\|v_{\mathrm{b}}\|^{2}/0.5^{2}\right)$ & 0.50 \\
Body Angular Velocity & $\exp\!\left(-\|\omega_{\mathrm{b}}\|^{2}/0.5\right)$ & 0.50 \\
Relative Body Position & $\exp\!\left(-\|p_{\mathrm{rel}}\|^{2}/0.1^{2}\right)$ & 1.00 \\
Action Rate & $\|a_t-a_{t-1}\|^{2}$ & $-0.01$ \\
Undesired Contacts & $\sum \mathbb{I}(F_c>1.0)$ & $-1.00$ \\
\bottomrule
\end{tabular}}
\end{minipage}
\hfill
\begin{minipage}{0.48\linewidth}
\centering
\caption{Domain Randomization parameters.}
\label{tab:domain_randomization}

\resizebox{\linewidth}{!}{
\begin{tabular}{llll}
\toprule
\textbf{Category} & \textbf{Parameter} & \textbf{Type} & \textbf{Range} \\
\midrule
\multirow{2}{*}{Env.}
 & Static Friction & Uniform & [0.6, 1.0] \\
 & Dynamic Friction & Uniform & [0.4, 0.8] \\
\midrule
Delays & Action Delay & Uniform & [0.0, 0.02] s \\
\midrule
\multirow{2}{*}{Motor}
 & Kp/Kd Factor & Scaling & [0.8, 1.2] \\
 & Joint Armature & Scaling & [0.8, 1.2] \\
\midrule
\multirow{3}{*}{Body}
 & Link Mass & Scaling & [0.8, 1.2] \\
 & Base Mass & Additive & [-5.0, 5.0] \\
 & Base CoM & Additive & [-0.05, 0.05] m \\
\bottomrule
\end{tabular}}
\end{minipage}

\end{table}

\end{document}